\documentclass[runningheads,a4paper]{llncs}

\usepackage{amssymb}
\usepackage{amsmath}
\usepackage{graphicx}
\usepackage{epsfig}
\usepackage{url}
\usepackage{subfigure}
\urldef{\mailsa}\path|{riccardo.satta, fumera, roli}@diee.unica.it|
\urldef{\mailsb}\path|{marco.cristani, vittorio.murino}@iit.it|
\newcommand{\keywords}[1]{\par\addvspace\baselineskip
\noindent\keywordname\enspace\ignorespaces#1}

\begin{document}

\mainmatter  

\title{A Multiple Component Matching Framework for Person Re-Identification}

\titlerunning{A Multiple Component Matching Framework for Person Re-Identification}

\author{Riccardo Satta%
\and Giorgio Fumera\and Fabio Roli\and Marco Cristani\and\\
Vittorio Murino}
\authorrunning{R. Satta, G. Fumera, F. Roli, M. Cristani and V. Murino}

\institute{Dept. of Electrical and Electronic Engineering, University of Cagliari\\
Piazza d'Armi, 09123 Cagliari, Italy\\
\mailsa\\
Istituto Italiano di Tecnologia (IIT)\\
Via Morego 30, 16163 Genova, Italy\\
\mailsb\\
}
\maketitle

\begin{abstract}
Person re-identification consists in recognizing an individual that has already been observed over a network of cameras. It is a novel and challenging research topic in computer vision, for which no reference framework exists yet. Despite this, previous works share similar representations of human body based on part decomposition and the implicit concept of multiple instances.
Building on these similarities, we propose a Multiple Component Matching (MCM) framework for the person re-identification problem, which is inspired by Multiple Component Learning, a framework recently proposed for object detection \cite{Dollar08-MCL}.
We show that previous techniques for person re-identification can be considered particular implementations of our MCM framework.
We then present a novel person re-identification technique as a direct, simple implementation of our framework, focused in particular on robustness to varying lighting conditions, and show that it can attain state of the art performances.
\end{abstract}

\section{Introduction}
\label{sec:intro}

In video surveillance, person re-identification is the task of recognizing an individual that has already been observed over a network of cameras. It is a novel research topic with many challenging issues, like low resolution frames, different and time-varying light conditions, and partial occlusions.

So far, no theoretical framework for the person re-identification problem exists yet. Previous works are based on different, apparently unrelated approaches, and are focused on devising effective features and appearance models.
Despite this, most works turn out to adopt a similar \textit{part-based} body representation, and/or use implicitly the concept of \textit{multiple instances} \cite{Yang05-MILsurvey} by considering image patches, regions, or points of interest.

It is worth to note that the above commonalities among previous works bear a resemblance with a framework recently proposed for object detection, Multiple Component Learning (MCL) \cite{Dollar08-MCL}. In fact, MCL adopts a part-based multiple-instance object representation: an object is considered as a sequence of parts, and each part is represented as a set of instances. 
A Multiple Instance Learning (MIL) approach is then used to recognise each individual part, using positive and negative examples of instances of that part.
Despite this analogy, the MCL framework can not be directly applied to person re-identification, which is a \emph{recognition} rather than a \emph{detection} task. Moreover, only one or a few positive examples (the template images of a given person) are usually available, while no negative examples are generally considered.
This makes person re-identification a task more suited to a matching approach rather than a recognition approach based on learning of human body models. In fact, most previous works formulated person re-identification as a task where template images of individuals are collected and then matched against probe images by some similarity measure.

Based on the above premises, in this work we propose a framework for person re-identification named Multiple Component Matching (MCM).
We embed in MCM the part-based multiple-instance body representation approach underlying both MCL and previous works on person re-identification, with the aim to provide a framework which can be used as a reference to develop new methods, and possibly to improve current ones.
We show at first that techniques proposed in previous works can be seen as particular implementations of MCM. Then, we present a novel person re-identification method, as a more direct and simple implementation of MCM, focused in particular on robustness to variations of lighting conditions, which attains state of the art performances.

We overview MCL and previous works on person re-identification in Sect.~\ref{sec:background}. The MCM framework is presented in Sect.~\ref{sec:mcm}, where its relationships with previous works are also discussed. Our implementation of MCM is presented in Sect.~\ref{sec:themethod}, and is experimentally evaluated in Sect.~\ref{sec:experiments}. In Sect.~\ref{sec:conclusions} future research directions are discussed.

\section{Background and Previous Works}
\label{sec:background}

Here we present first the MCL framework, and then overview previous works on person re-identification.

\subsection{Multiple Component Learning}
\label{subsec:mcl}

Multiple instance learning (MIL) is a general learning paradigm for problems in which samples are made up by a \emph{bag} (set) of labeled \emph{instances}, and only the label of the whole bag is known. The task is to build a classifier which learns to label bags using the feature vectors of their instances \cite{Yang05-MILsurvey}.
MIL has been applied to several computer vision problems, including scene classification \cite{Maron98}, image retrieval \cite{ZhangD10}, and object detection \cite{Viola05}.

MCL is an extension of MIL, tailored to object detection \cite{Dollar08-MCL}.
In MCL an object is represented as a \textit{sequence of sets of components} (bags of instances in MIL terminology).
The rationale behind is the independent detection of different object components, to gain robustness to partial occlusions.
Moreover, the subdivision in components is not predefined, but obtained as part of the learning process.
To this aim, the image is first subdivided into a predefined set of randomly chosen regions, and a classifier is trained on each region to detect the corresponding component.
Since the corresponding object component may appear in different positions inside a region, each region is randomly subdivided into a set of possibly overlapping subregions (\emph{patches}), only some of which may contain the component.
A MIL classifier is then used to detect the corresponding component, threating each patch as a single instance.
An ensemble of MIL classifiers, one for each region, is trained via a boosting algorithm, so that the most discriminative regions get a higher weight.
Learning is thus done both at the \textit{set} level (to detect individual object components) and at the \textit{sequence} level (to combine the information on object components coming from different regions).
Note that, according to the MIL paradigm, in the training set only the image regions are labeled, while the patches inside each region are not.

\subsection{Previous Works on Person Re-identification}
\label{subsec:personreid}

Person re-identification consists in associating an individual from a probe set to the corresponding template in a gallery set.
Depending on the number of available frames per individual, the following scenarios can be defined \cite{Farenzena2010}: \textit{Single vs Single} (SvsS), if only one frame per individual is available both in probe and in gallery sets; \textit{Multiple vs Single} (MvsS), if multiple template frames per individual are available in the gallery set; \textit{Multiple vs Multiple} (MvsM), if multiple frames per individual are available both in the probe and gallery sets.
The most challenging scenario is SvsS.

All the above scenarios have been considered in \cite{Farenzena2010}. Here, human body is subdivided with respect to its symmetry properties: anti-symmetry separates head, torso and legs, while symmetry is exploited to divide left and right parts. 
The descriptor is made up of three local features: colour histograms for torso and legs, weighted with respect to the distance from the symmetry axis; \textit{maximally stable colour regions} (MSCR) and \textit{recurrent high-structured patches} (RHSP), both extracted from torso and legs separately.
To obtain MSCR and RHSP, several patches are sampled at random, mainly near symmetry axes; then, clustering algorithms are used to find the most significant ones.
The matching distance is a combination of the distances computed on the individual features. In MvsS and MvsM scenarios, templates are accumulated into a single descriptor.

In \cite{Bak2010}, an human body parts detector is used to find in the body of each individual fifteen non-overlapping square cells, that have proven to be ``stable regions'' of the silhouette. For each cell a covariance descriptor based on colour gradients is computed. Descriptor generation and matching is performed through a pyramid matching kernel. This method can be applied only to SvsS scenarios.

In \cite{Bak2010-2} two methods were proposed.
In the first, Haar-like features are extracted from the whole body, while in the second the body is divided into upper and lower part, each described by the MPEG7 Dominant Colour descriptor.
Learning is performed in both methods, respectively to choose the best features and to find the most discriminative appearance model.
The training set for each individual consists of different frames as positive examples (MvsS, MvsM scenarios), and of everything which is not the object of interest as negative examples.
In the SvsS scenario (one frame per individual), different viewpoints are obtained by sliding a window over the image in different directions.

An approach based on harvesting SIFT-like interest points from different frames of a video sequence is described in \cite{Hamdoun2008}.
Different frames are used also in \cite{Gheissari2006}, where two methods are proposed. The first one is based on interest points, selected on each frame by the Hessian-Affine interest operator. The second one exploits a part subdivision of the human body based on decomposable triangulated graphs and dynamic programming to find the optimal deformation of this model for the different individuals. Each part is then described by  features based on colour and shape; the distance between a template and a probe is a combination of the distances between pairs of corresponding parts.

In \cite{Gray2008} the problem of defining the best descriptor for person re-identification is addressed. Different features are extracted, and their weights are computed by a boosting algorithm.
Features are computed from randomly taken strips.

In \cite{ProsserReIdBMVC2010} person re-identification is considered as a relative ranking problem, exploiting a discriminative subspace built by means of an Ensemble RankSVM. Colour and texture-based features are extracted from six fixed horizontal regions.

Despite the methods summarised above exhibit many differences, it can be noted that all of them are based on some part-based body representation, and/or exploit more or less implicitly the concept of multiple instances. This provides the foundation for the proposed framework, which is depicted in the next section.

\section{A Framework for Person Re-identification}
\label{sec:mcm}

In this section we describe the proposed Multiple Component Matching (MCM) framework for person re-identification.
As mentioned previously, MCM is inspired by MCL; in fact, we found that the concepts behind most previous work are similar to the ones underlying MCL, namely part subdivision and multiple component representation.

Like in MCL, an object is represented as an ordered sequence of sets. In turn, each set is made up by several components.
Differently from object detection problems addressed by MCL, we view person re-identification as an object recognition problem where a matching approach is used without any learning phase: while in the training samples of MCL a set is composed by both negative and positive components (the first contain the object part of interest, while the latter do not), in MCM only positive ones are available, namely only those corresponding to body parts of the template person.
\begin{figure}[t]
\begin{center}
   \includegraphics[trim = 0mm 0mm 10mm 0mm, clip, width=0.7\linewidth]{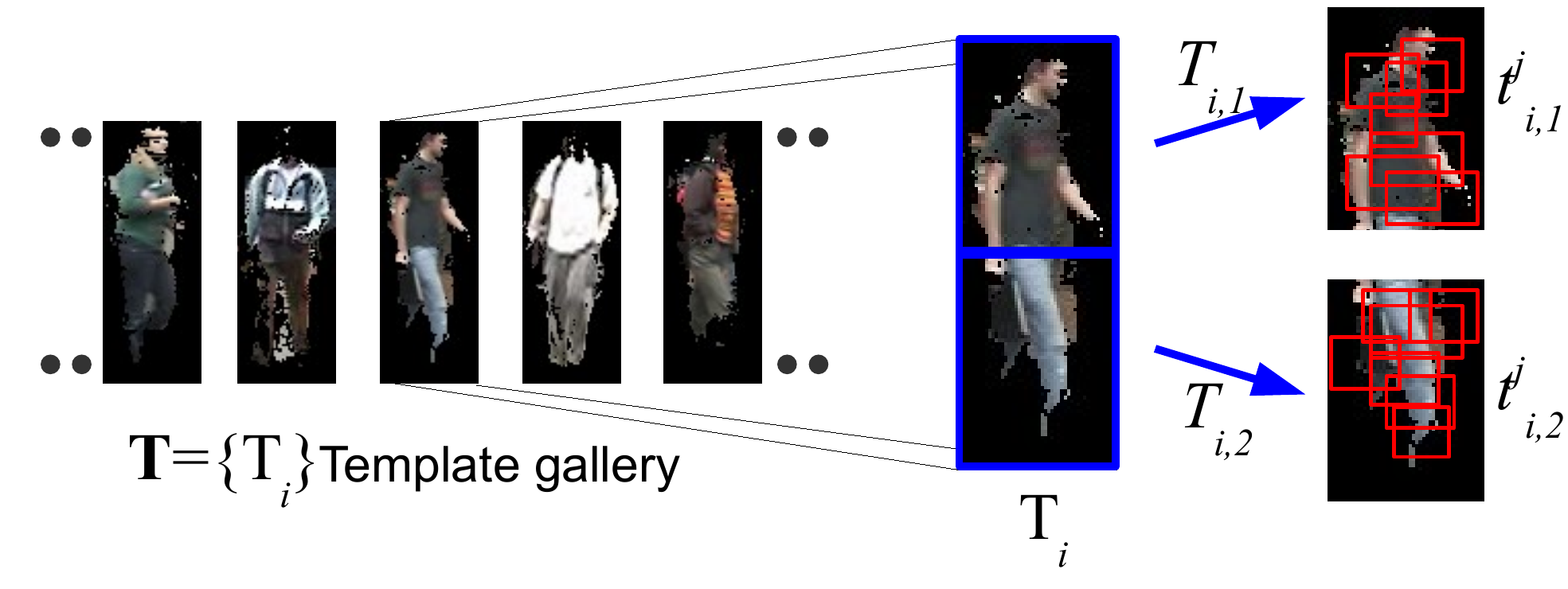}
\end{center}
   \caption{Representation of an individual according to MCM. Each template $\mathbf{T}_i$ of the gallery $\mathcal{T}$ is represented by an ordered sequence of $M$ parts $T_{i,j}$ (here $M=2$, corresponding to upper and lower body parts). 
Each part is made up of a set of components (here, rectangular patches, in red). A feature vector $t^k_{i,j}$ describes each component.}
\label{fig:MCM}
\end{figure}
Formally, let $\mathcal{T}=\{\mathbf{T}_1,\ldots,\mathbf{T}_N\}$ be the \textit{gallery set} of templates, each corresponding to an individual.
Each template $\mathbf{T}_i$ is represented by an ordered sequence of a predefined number of $M$ sets, corresponding to the $M$ parts into which an image is subdivided:
\begin{equation}
\label{eq:MCMTemplateGallery}
 \mathbf{T}_i = \{T_{i,1},\ldots,T_{i,M}\}
\end{equation}
Following a multiple-instance representation, every part $T_{i,j}$ is represented by a set of an arbitrary number $n_{i,j}$ of elements (instances in MIL, components in MCL) (see Fig.~\ref{fig:MCM}), and is described by the corresponding feature vectors $\mathbf t^k_{i,j}$:
\begin{equation}
\label{eq:MCMTemplateSet}
T_{i,j}=\{\mathbf t^1_{i,j},\ldots,\mathbf t^{n_{i,j}}_{i,j}\}, \mathbf t^k_{i,j}\in{\mathbb{X}} ,
\end{equation}
where $\mathbb{X}$ denotes the feature space (assumed the same for all sets, for the sake of simplicity, and without losing generality).
Given a probe $\mathbf{Q}$, which is represented as a sequence of parts as described above, the task of MCM is to find the most similar template $\mathbf{T}^*\in\mathcal{T}$, with respect to a similarity measure $D(\cdot,\cdot)$:
\begin{equation}
\label{eq:MCMSetMatching}
 \mathbf{T}^* = \operatorname*{arg\,min}_{\mathbf{T}_i} D(\mathbf{T}_i ,\mathbf{Q}) .
\end{equation}
We consider a similarity measure $D$ between sequences defined as a combination of similarity measures $d(\cdot,\cdot)$ between sets:
\begin{equation}
\label{eq:MCMSequenceMatching}
 D(\mathbf{T}_i,\mathbf{Q}) = f\big{(}d(T_{i,1}, Q_1),\ldots,d(T_{i,M}, Q_M)\big{)} .
\end{equation}

Similarly to MCL, where learning is performed both at the sequence level and at the set level, in MCM the two similarity measures $D$ and $d$ are defined at sequence and at the set level.
The similarity measure between sequences $D$ can be any combination of the set distances, like a weighted average in which the coefficients reflect the relevance of the corresponding regions.
The following considerations can be made on the choice of a proper similarity measure $d$ between sets. In MCL, this level corresponds to build a MIL classifier.
In the MIL paradigm, only a subset of the instances belonging to a set may be responsible of the label of the whole set. Analogously, in MCM a template set can be considered to match the corresponding probe set, if at least a few pairs of instances of the two sets are ``similar'': the object components which may be in common to the two parts represented by the sets can be placed, in fact, everywhere inside the parts, and therefore be ``captured'' by any of the instances.
  
Accordingly, the similarity measure between sets can be defined as the minimum of the similarity measures between all pairs of their instances. To minimize the sensitivity to outliers, more complex measures can be defined, for example considering the first $k$ best matches instead of only one. Moreover, the similarity measure can take into account relationships among instances in a set, i.e.~the relative spatial disposition of the corresponding components.

\subsection*{Previous Works and MCM}
\label{sec:analogies}

Most of the previous works on person re-identification described in Sect.~\ref{subsec:personreid} can be framed into MCM.

A common way to face the kinematics of the human body is to adopt a division in parts \cite{Bak2010-2,Bak2010,Farenzena2010,Gheissari2006,ProsserReIdBMVC2010}, consistently with the part subdivision of MCM. In MCM, every part is made up of several components. In \cite{Farenzena2010}, two of the three features proposed are based on such subdivision: both MSCR and RHSP represent multiple patches of the considered body part. When parts are not represented by components, as in the third feature of \cite{Farenzena2010}, and in \cite{Bak2010-2,Bak2010,Gheissari2006,ProsserReIdBMVC2010}, this can still be considered as a special case of MCM, where every part is composed by only one component.
On the contrary, while in \cite{Bak2010-2,Gheissari2006,Gray2008,Hamdoun2008} no part-based body representation is used, a multiple-instance representation is nevertheless adopted, in the form of interest points or patches. These methods can be seen as particular implementations of MCM as well, where only one body part is considered.

In the definition of MCM, we assumed that every part is represented in the same feature space. This is however not a strict requirement.
If different feature vector representations are used, a part can be represented by several feature sets.
Accordingly, also the methods in \cite{Farenzena2010,Gray2008}, where several feature vector representations were used, can be seen as MCM implementations.

\section{A Method for Person Re-identification Based on MCM}\label{sec:themethod}

We propose here a novel person re-identification method as a possible example of a direct implementation of the MCM framework.
In particular, our method exploits MCM to attain robustness to changing illumination conditions.

\begin{figure}[t]
\begin{center}
   \includegraphics[trim = 0mm 95mm 0mm 0mm, clip, width=0.32\linewidth]{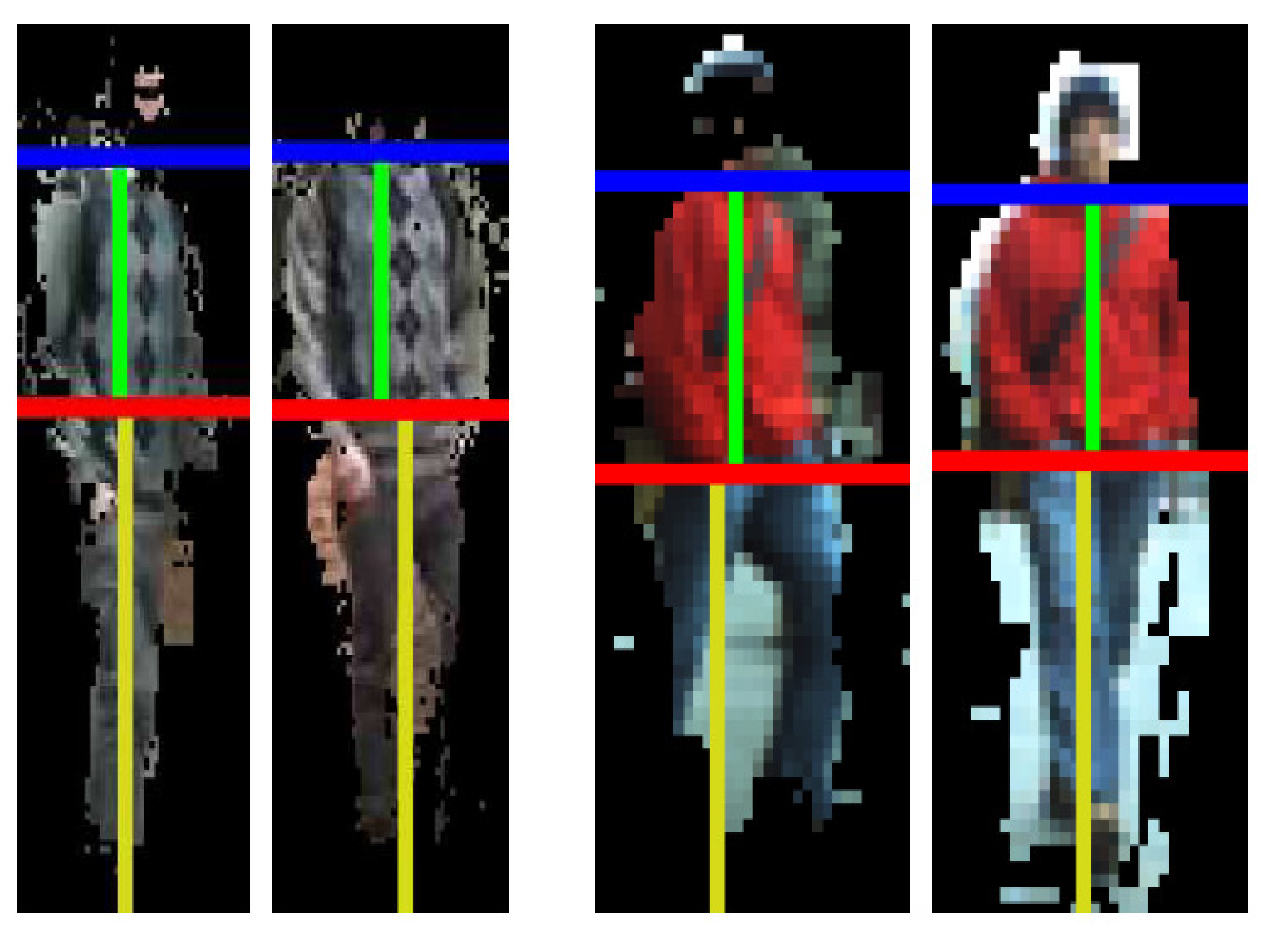}
\end{center}
   \caption{Body partition: symmetry (vertical) and anti-simmetry (horizontal) axes.}
\label{fig:bodyPartition}
\end{figure}

We assume that the ``blob'' of the person has been already extracted by some detector, and thus only pixels belonging to the mask of the blob are considered.
To divide the body into parts, we exploit the anti-symmetry axes as proposed in \cite{Farenzena2010} (see Fig.~\ref{fig:bodyPartition}) to locate torso and legs. The head is discarded, since it does not carry enough information due to its relatively small size.
Every body part is represented as a set of a fixed number $P$ of rectangular patches of random area in the range $[12.5\%,25\%]$ of the region area.
Every patch is described by a pair $(HSV,y_{pos})$, where $HSV$ is the concatenation of H, S and V histograms of the patch (24, 12 and 4 bins respectively) and $y_{pos}$ is the relative vertical position of the center of the patch, with respect to the height of the region.

As stated in Sect.~\ref{sec:mcm}, a challenging issue in person re-identification is how to face lighting changes. MCM suggests a way: the multiple instance representation, in fact, can naturally be exploited here, adding instances corresponding to different lighting conditions. However, such instances can not easily been obtained from real data (multiple frames, corresponding to as much as possible different illuminations, including lighting gradients and shadows, should be acquired). So, we \emph{simulate} them by constructing artificial patches from real template ones.

\begin{figure}[t]
\begin{center}
  \subfigure[]
  {\includegraphics[width=0.34\linewidth]{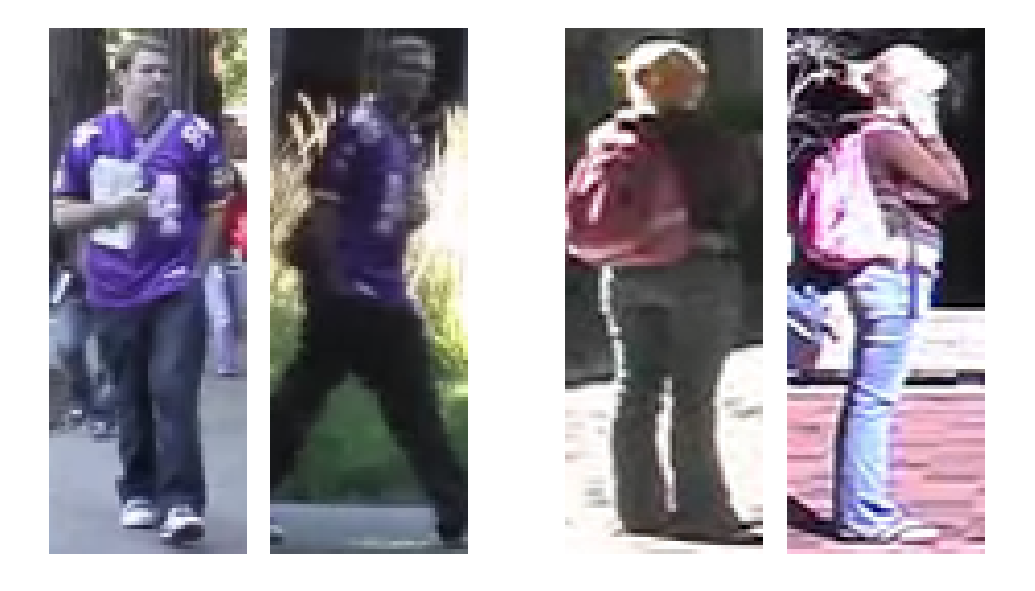}}
\hspace{2.5mm}
  \subfigure[]
  {\includegraphics[width=0.43\linewidth, clip=true, trim=0mm 7mm 0mm 0mm]{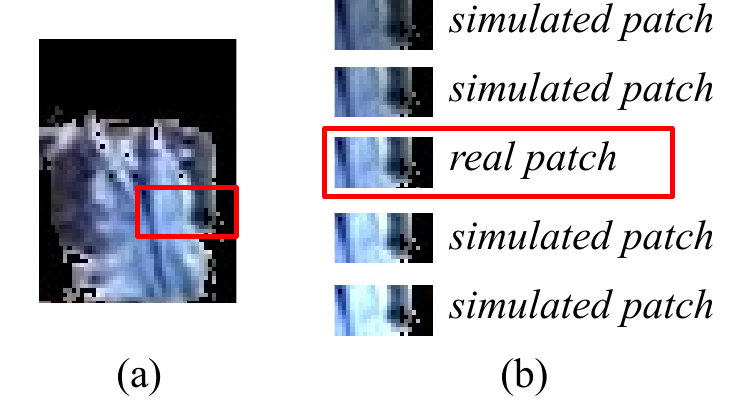}}
\end{center}
   \caption{(a) Two examples of different images for the same individual: note the difference both in contrast and brightness. (b) Examples of four artificial patches simulating changing illumination (right), corresponding to the patch highlighted on the left.}
\label{fig:differentLightConditions}
\end{figure}

Light variations usually result in a change of both brightness and contrast of the image (see for example Fig.~\ref{fig:differentLightConditions}-a). Brightness variations can be obtained by adding or subtracting a fixed value to the RGB components of the pixels of the image. Instead, changing contrast means increasing or decreasing the differences between pixel values. A standard method to obtain this is the following: denoting as $[0, C]$ the original range of each colour channel (usually $C=255$), every R, G, and B pixel value is translated to $[-C/2,C/2]$, multiplied by a fixed coefficient, and then re-normalised to $[0,C]$. A coefficient greater than 1 results in a higher contrast, while a lower contrast is obtained by choosing values smaller than 1.

To change both brightness and contrast, we propose a modification of the above method, which does not translate values to $[-C/2,C/2]$ first, but simply multiplies each pixel value of each channel by a coefficient $K$.
Intuitively, this increases (or decreases) the differences between pixel values as well. However, while in the standard method values lower than $C/2$ are reduced, and those higher than $C/2$ are increased, in our variant all the values are increased (or decreased), thus obtaining also a change of brightness.
Our algorithm multiplies pixel values by a series of coefficients $[k_1,\ldots,k_S]$ to generate $S$ simulated patches from each real one (see the example in Fig.~\ref{fig:differentLightConditions}-b).
To choose proper $k_i$ values, we start from an initial vector $K = [k_{1},\ldots,k_{S}]$, then decrease its values until applying the greatest $k_i$ to the original image does not saturate the image too much. More precisely, we check that the mean value of R, G and B multiplied by the greatest value of $K$ is not higher than a threshold, which we set to 240.

As explained in Sect.~\ref{sec:mcm}, to implement MCM two similarity measures $D$ and $d$ have to be defined.
$D$ was defined as the average of the distances between sets. Concerning $d$, in set theory a common distance function between sets is the \textit{Hausdorff Distance} \cite{Edgar1990}, defined as the maximum of the minimum distances between each element of one set and each element of the other. Comparing two sets $X=\{x_i\}$ and $Y=\{y_i\}$, we have
\begin{equation}\label{eq:haussdistance}
d_{H}(X, Y) = \max(h(X, Y),h(Y, X))
\end{equation}
where
\begin{equation}
 h(X, Y) = \max_{x\in{X}}\min_{y\in{Y}}(\| x - y\|)
\end{equation}
Such a distance measure is sensitive to outlying elements. To avoid this issue, we adopted the \textit{k-th Hausdorff Distance} proposed by Wang and Zucker \cite{Wang2000}, which takes the $k$-th ranked distance rather than the maximum: in Eq. ~\ref{eq:haussdistance}, in place of $h(X, Y)$ we have then
\begin{equation}\label{eq:kthhauss}
 h_k(X, Y) = \operatorname*{\textit{kth}\,}_{x\in{X}}\min_{y\in{Y}}(\| x - y\|)
\end{equation}
Finally, to compute the norm $\| x - y\|$ in Eq.~\ref{eq:kthhauss}, a distance metric must be defined for the pairs $(HSV,y_{pos})$ that describe each patch.
Denoting with $b(HSV_1,HSV_2)$ the Bhattacharyya distance between histograms, we defined the metric as
\begin{multline*}
(HSV_1,y_{pos,1}) - (HSV_2,y_{pos,2}) = b(HSV_1,HSV_2)\cdot(1+\beta)|y_{pos,1} - y_{pos,2}|
\end{multline*} 
where $\beta$ controls the relevance of the difference in spatial position of the patches.

The above method is applicable only to SvsS scenarios. To extend it to MvsS and MvsM scenarios, one possible approach is to accumulate the patches over different frames in the same sequence of sets.

\section{Experimental Results}
\label{sec:experiments}

The performance of the proposed MCM implementation was assessed on the VIPeR benchmark dataset \cite{Gray2007evaluating}, a challenging corpus composed by two non overlapping views of 632 different pedestrians, which show varying changing conditions and pose variations. The best performing method so far is SDALF \cite{Farenzena2010}.

We evaluated performance in terms of cumulative matching characteristics (CMC) curve, which represents the probability of finding the correct match over the first $n$ ranks. As in \cite{Farenzena2010}, we obtained blob masks by the STEL generative model \cite{Jojic2009}.
In our method we set $\beta=0.6$, used $P=80$ real patches, and adopted $K = [1.4, 1.2, 1.0, 0.8, 0.6]$ as the initial vector coefficients for simulation. The value of $k$ for the $k$-th Hausdorff Distance measure was set to $10$.

We employed the same experimental setup of \cite{Farenzena2010}, to obtain comparable results. Ten random subsets of 316 pedestrians were drawn from the original dataset. The gallery set is composed by the first image of each person; the probe set, by the second one. Images of the probe set are compared to the images of the gallery set to find the best match. We used the same images as in \cite{Farenzena2010}.
 
\begin{figure*}[t]
\begin{center}
   \includegraphics[width=0.483\linewidth, clip=true, trim=11mm 3mm 25mm 2mm]{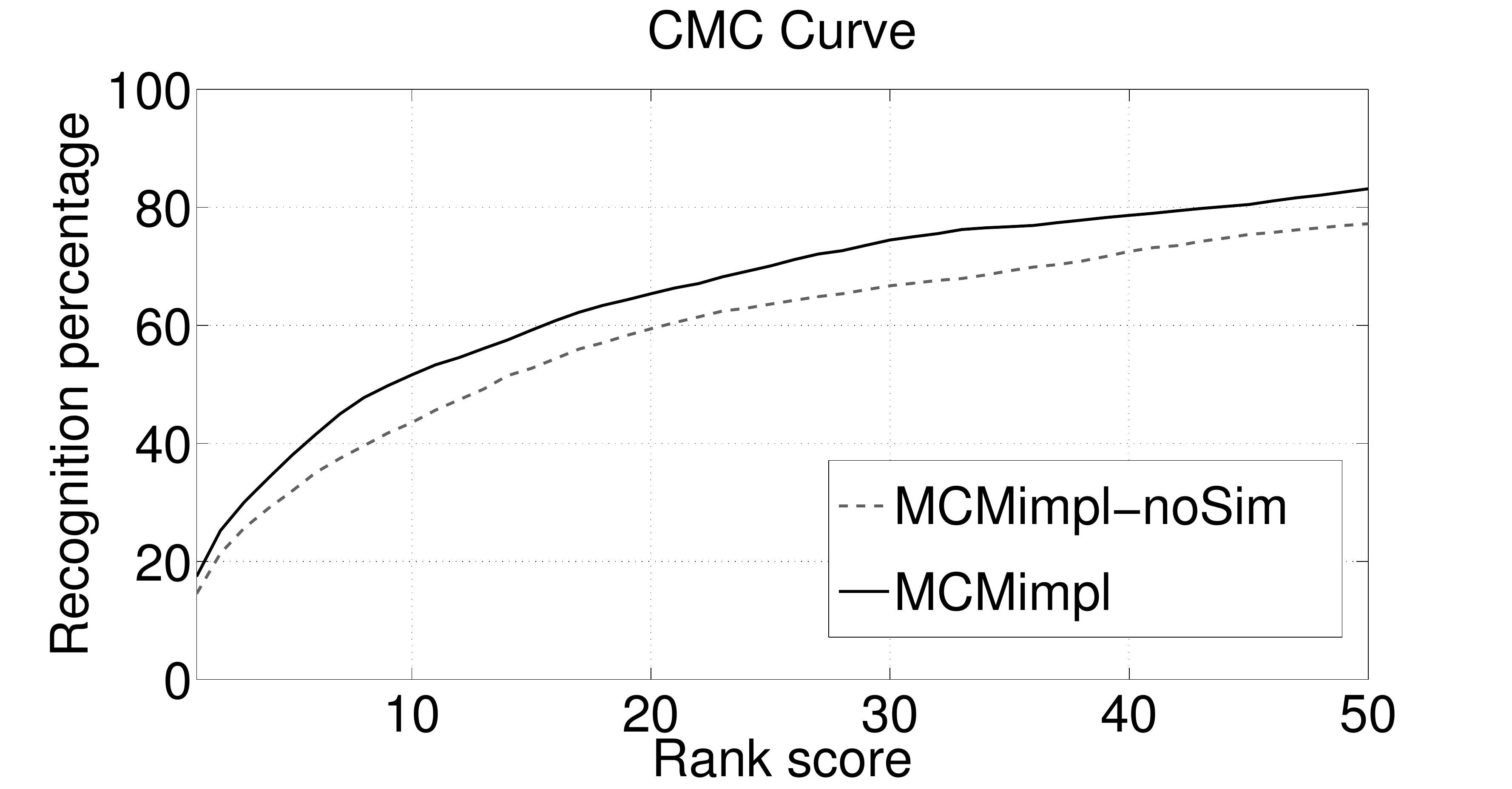}
   \includegraphics[width=0.483\linewidth, clip=true, trim=8mm 3mm 28mm 2mm]{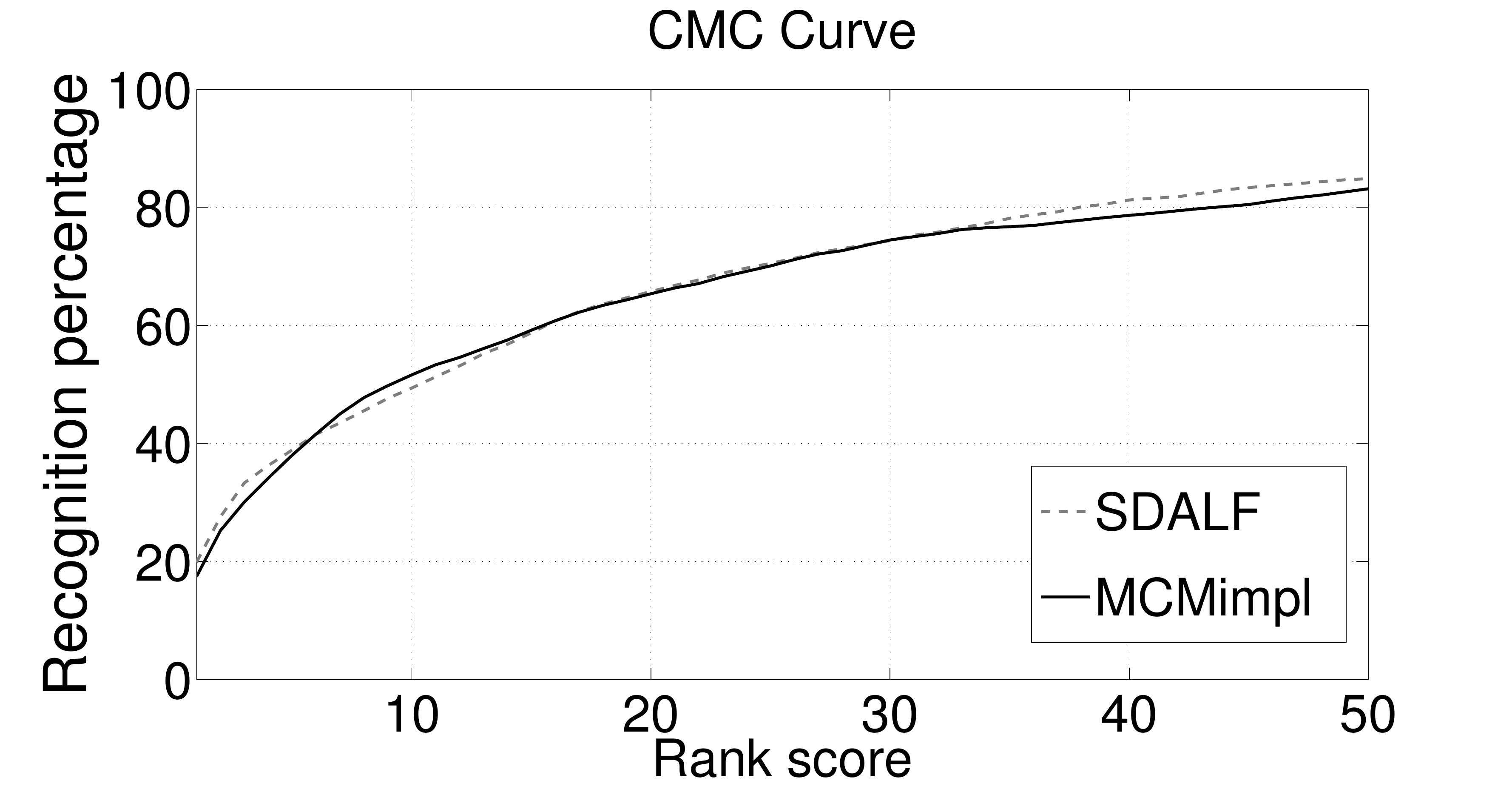}
\end{center}
   \caption{Performances on the ViPER dataset. (left) Our MCM implementation (\textit{MCMimpl}) with and without simulation. (right) \textit{MCMimpl} compared with \textit{SDALF}.}
\label{fig:CMCviper}
\end{figure*}

In Fig.~\ref{fig:CMCviper}(left), we reported the CMC curve attained by our method with and without simulation.
Simulating varying lighting conditions allows to attain a much better performance, at the expense of a higher overall computation time which is due to template creation and to the higher number of patches compared.

\begin{table*}[b]
\centering
\begin{tabular}{l||c|c|c}
                    & Template desc. creation   & Probe desc. creation  & Matching \\
\hline
$MCM_{impl}$        & $93.7 ms$                 & $6.8 ms$              & $28.6 ms$ \\
$MCM_{impl,nosim}$  & $ 6.8 ms$                 & $6.8 ms$              & $6.7 ms$ \\
\hline
\end{tabular}
\caption{Average computation time per frame or, for matching, per pair of frames.}
\label{tab:computationTimes}
\end{table*}
In Fig.~\ref{fig:CMCviper}(right) we compare the performance of our method, including the simulation, with SDALF. As shown, the proposed implementation of MCM attains a performance which is close to the reference. What make $MCMimpl$ highly preferable is the computational cost: SDALF involves time-consuming operations like clustering, transformations, cross-correlation, etc., while $MCMimpl$ performs only simple and fast operations.

The average computation time of our proposed method on a 2.4 GHz CPU is reported in Tab.~\ref{tab:computationTimes}. The method was implemented in C++, without any particular optimization or parallelization. As a qualitative comparison, the implementation of SDALF made available by the authors of SDALF, written partly in C++ and partly in MATLAB, requires over 13 seconds to build one descriptor, and performs a match in around 60 ms, on the same 2.4 GHz CPU. Note that these computational times can not be directly compared to the ones reported for $MCMimpl$, due to the partial MATLAB implementation.
However, reimplementing MATLAB code in a more performing language usually results in a speed-up of no more than 10-20 times, pretty far from the difference of 2 orders of magnitude between the descriptor creation times.

\section{Conclusions and Future Work}
\label{sec:conclusions}

We proposed a framework for person re-identification which embeds common ideas underlying most of the previous works, and is inspired by the MCL framework for object detection.
We also developed a simple MCM implementation including a method to make it robust to changing illumination conditions, which has a low computational cost and attains a performance close to the state-of-the-art method on a benchmark data set.

Two main directions for further research can be foreseen.
First, simulation can be implemented in MCM to attain robustness also to pose variations.
Second, it is interesting to investigate whether MCM can be extended to a learning approach. Exploiting its relationships with MCL, this can enable MCM to adopt a MIL approach to learn the appearance of each body part, as an alternative to the matching approach considered here, taking advance of the large available literature on that learning framework.

\bibliographystyle{splncs}
\bibliography{egbib}
\end{document}